\documentclass[conference]{IEEEtran}
\IEEEoverridecommandlockouts
\usepackage{cite}
\usepackage{amsmath,amssymb,amsfonts}
\usepackage{algorithmic}
\usepackage{graphicx}
\usepackage{threeparttable}
\usepackage{textcomp}
\usepackage{xcolor}

\usepackage{algorithm}

\usepackage{booktabs} 
\usepackage{caption}  

\usepackage{multirow} 
\usepackage{amssymb}  

\usepackage{pifont}
\usepackage{array}

\usepackage{subcaption}

\usepackage{geometry}
\def\BibTeX{{\rm B\kern-.05em{\sc i\kern-.025em b}\kern-.08em
    T\kern-.1667em\lower.7ex\hbox{E}\kern-.125emX}}
\usepackage{caption}
\begin{document}

\title{A Memory-Efficient Framework for Deformable Transformer with Neural Architecture Search}

\author{
    \IEEEauthorblockN{{\color{black}Wendong Mao\textsuperscript{$a$}, Mingfan Zhao\textsuperscript{$a$}, Jianfeng Guan\textsuperscript{$a$}, Qiwei Dong\textsuperscript{$b$}, Zhongfeng Wang\textsuperscript{$a$}}}
    \IEEEauthorblockA{{\color{black}\textsuperscript{$a$}School of Integrated Circuits, Sun Yat-Sen University, Shenzhen, China}}
    \IEEEauthorblockA{{\color{black}\textsuperscript{$b$}School of Electronic Science and Engineering, Nanjing University, Nanjing, China}}
    
    \IEEEauthorblockA{{\color{black}Email: maowd@mail.sysu.edu.cn, zhaomf7@mail2.sysu.edu.cn, guanjf@mail2.sysu.edu.cn}}
    \IEEEauthorblockA{{\color{black} qiweidong@smail.nju.edu.cn, wangzf83@mail.sysu.edu.cn}}    
    }

\maketitle

\begin{abstract}
Deformable Attention Transformers (DAT) have shown remarkable performance in computer vision tasks by adaptively focusing on informative image regions. However, their data-dependent sampling mechanism introduces irregular memory access patterns, posing significant challenges for efficient hardware deployment. Existing acceleration methods either incur high hardware overhead or compromise model accuracy. To address these issues, this paper proposes a hardware-friendly optimization framework for DAT. First, a neural architecture search (NAS)-based method with a new slicing strategy is proposed to automatically divide the input feature into uniform patches during the inference process, avoiding memory conflicts without modifying model architecture.  The method explores the optimal slice configuration by jointly optimizing hardware cost and inference accuracy. Secondly, an FPGA-based verification system is designed to test the performance of this framework on edge-side hardware. Algorithm experiments on the ImageNet-1K dataset demonstrate that our hardware-friendly framework can maintain have only 0.2\% accuracy drop compared to the baseline DAT. Hardware experiments on Xilinx FPGA show the proposed method reduces DRAM access times to 18\% compared with existing DAT acceleration methods.
\end{abstract}


\section{Introduction}
Nowadays, Transformer~\cite{1} has shown outstanding performance in natural language processing (NLP). As the potential of Transformers became evident, researchers extended their application to computer vision (CV), leading to the development of the Vision Transformer (ViT) architecture~\cite{2}. ViT has achieved impressive results across various CV tasks, such as object detection~\cite{3}, image classification~\cite{4}, and image segmentation~\cite{5}. Based on ViT's self-attention mechanism, numerous subsequent works~\cite{6,Mobilevit,LeViT} have been proposed to further enhance the performance and efficiency of transformers on visual tasks. However, Xia et al.\cite{7} pointed out that when too many keys correspond to a single query in visual Transformers, it can lead to high computational cost, slow convergence, and an increased risk of overfitting. This has inspired the emergence of new attention mechanisms that enable the key/value set for a given query to be both flexible and adaptive.

\begin{figure}[htb]
  \centering
  \includegraphics[width=\linewidth]{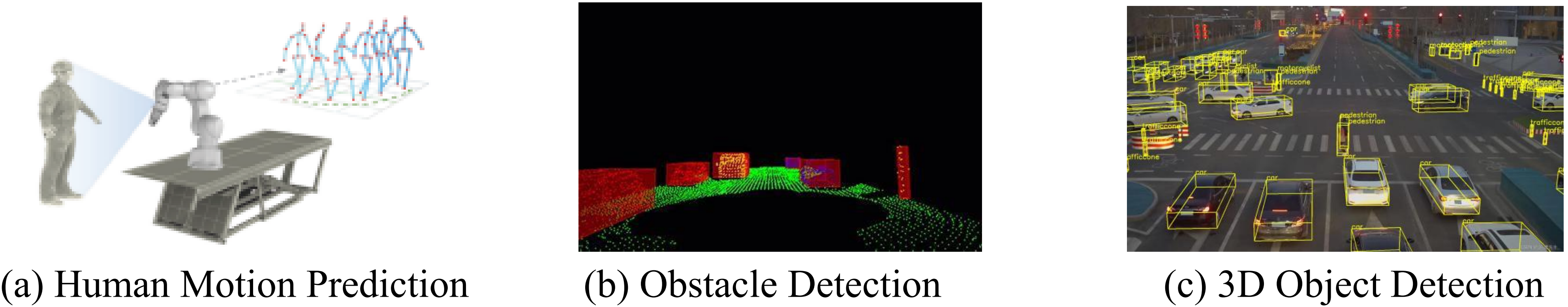}
  \caption{Practical applications of deformable attention.}
  \label{fig-1}
  \vspace{-3mm}
\end{figure}

The development of Deformable Attention Transformer (DAT)~\cite{7} effectively solves the above issue.
In its deformable attention mechanism, the sampling positions of key-value pairs are data-dependent rather than fixed. Consequently, the self-attention module can more effectively focus on important image regions that contain informative features. This flexible attention mechanism facilitates DAT's application across diverse domains such as robotics and autonomous driving, as illustrated in Fig.~\ref{fig-1}.

However, deploying the DAT on edge platforms, such as intelligent vehicles and mobile devices, remains challenging due to the data-dependent nature of its deformable attention, which dynamically samples keys and values from feature maps. This data dependency induces random and conflicting memory accesses. Therefore, traditional Transformer accelerators like SpAtten~\cite{8} and ELSA~\cite{9} are not well-suited for accelerating deformable Transformers because of their unique architectural characteristics.
DEFA~\cite{10} is a pioneering work to accelerate multi-scale deformable attention, which introduces pruning-assisted sampling to reduce the memory footprint of feature map sampling. 
Nevertheless, DEFA suffers from considerable hardware overhead. To address this limitation, this paper proposes a hardware-friendly and resource-efficient acceleration method for deformable attention.

In this paper, we propose a slicing-based acceleration strategy for the deformable attention mechanism, along with an optimal slice size search algorithm. The slicing strategy enables efficient computation of deformable attention on hardware platforms with limited resources. Furthermore, we introduce a neural architecture search (NAS)-based algorithm to determine the optimal slicing configuration.

The main contributions of this work are as follows:

\begin{itemize}
\item We propose a training-free slicing method, which does not change training process and only divides the input image into local patches during inference with the pre-established strategy. It decoupled the serial dependencies among different input tiles, reducing memory requirements without changing model architecture.
\item We develop a memory-aware NAS algorithm to construct a continuous search space encompassing various slicing strategies. It can determine the optimal slicing configuration, achieving the optimal balance between accuracy and hardware overhead. 
\item We deploy the proposed framework on the Xilinx FPGA platform, and experimental results demonstrate its multiple advantages in terms of hardware overhead and algorithm accuracy.
\end{itemize}

\section{A Training-Free Slicing Method for Deformable Attention}


Conventional self-attention modules acquire key and value information uniformly and sequentially across the image, enabling relatively straightforward application of acceleration techniques like parallel computation. In contrast, deformable attention mechanisms assign each reference point a learned random offset to determine its sampling position. This leads to completely random access patterns to the input image, resulting in the following challenges:




\subsubsection{Memory access conflicts} Multiple reference points may be offset to nearby image locations and access identical sampling point information, causing simultaneous access to the same memory location.

\subsubsection{Large memory overhead} Random access to the input image makes it necessary to store the entire input image in a large buffer for computing the sampled features. Meanwhile, the computation must be performed on the full image simultaneously, preventing dividing the image into independent regions for parallel processing, which results in significant memory and hardware resource overhead. 

\subsubsection{Serial processing dependency} Disordered memory access impedes parallel computing strategies and requires serial computations, which leads to low computational efficiency.

\begin{figure}[htb]
  \centering
  \includegraphics[width=0.95\linewidth]{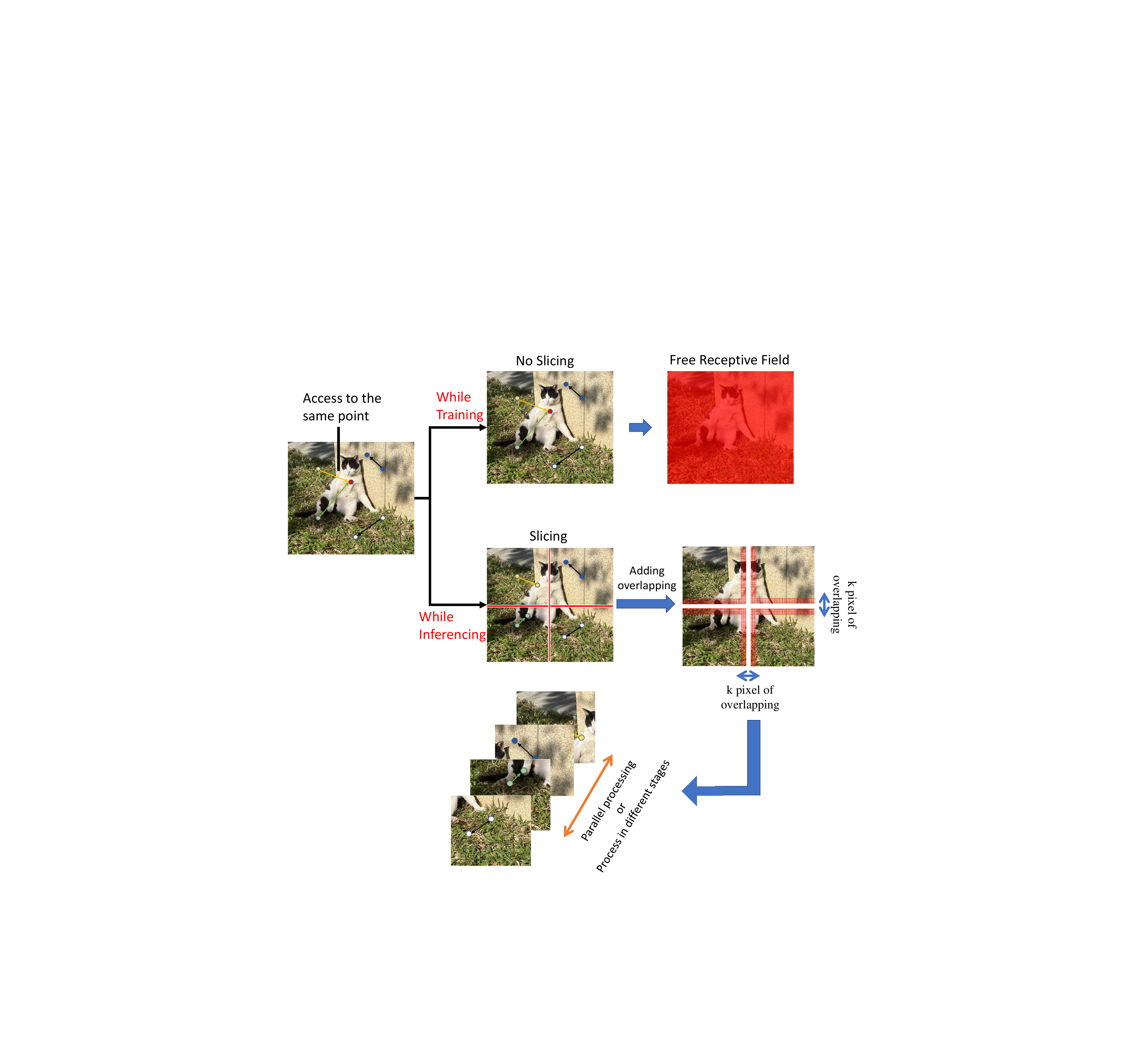}
  \caption{Illustration of the proposed slicing strategy.}
  \label{fig-3}
\end{figure}

To address the aforementioned issues, we propose a traing-free slicing strategy, as illustrated in Fig.~\ref{fig-3}. Specifically, during inference, the input image is divided into several independent patches, while training process remains unchanged. Reference points in a given patch are constrained so that they cannot shift outside their own patch, and no attention is computed across patches. After slicing, each patch can be processed individually in separate stages, which greatly reduces hardware resource consumption. As the patch size decreases, the required hardware resources drop exponentially. This is particularly significant for deploying deformable transformers on resource-constrained platforms such as FPGAs and mobile devices. Moreover, when hardware resources permit, computations on different patches can be parallelized, further improving computational efficiency.

However, if a reference point in the original image is shifted beyond its patch after slicing, the constrained offset range may lead to a decrease in model accuracy. To mitigate this issue, we introduce an overlapping region of width \textit{k} at the patch boundaries, as shown in Fig.~\ref{fig-3}. This overlapping region contains edge information from adjacent patches, which helps preserve model accuracy by alleviating the impact of restricted reference point movement.



\section{Neural Architecture Search for Optimal Slicing Strategy}

\begin{figure}[htb]
  \centering
  \begin{subfigure}[b]{\linewidth}
    \includegraphics[width=\linewidth]{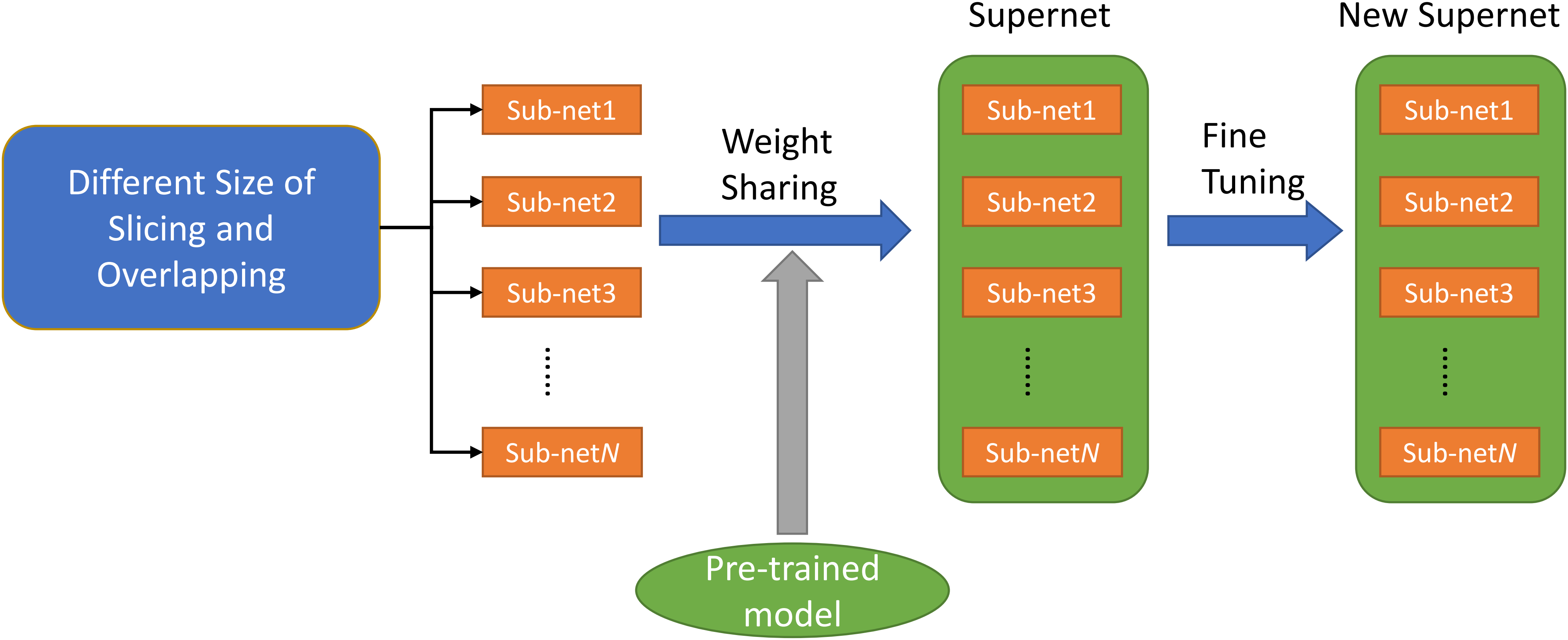}
    \caption{ Supernet Construction}
    \label{fig7a}
  \end{subfigure}
  \begin{subfigure}[b]{\linewidth}
    \includegraphics[width=\linewidth]{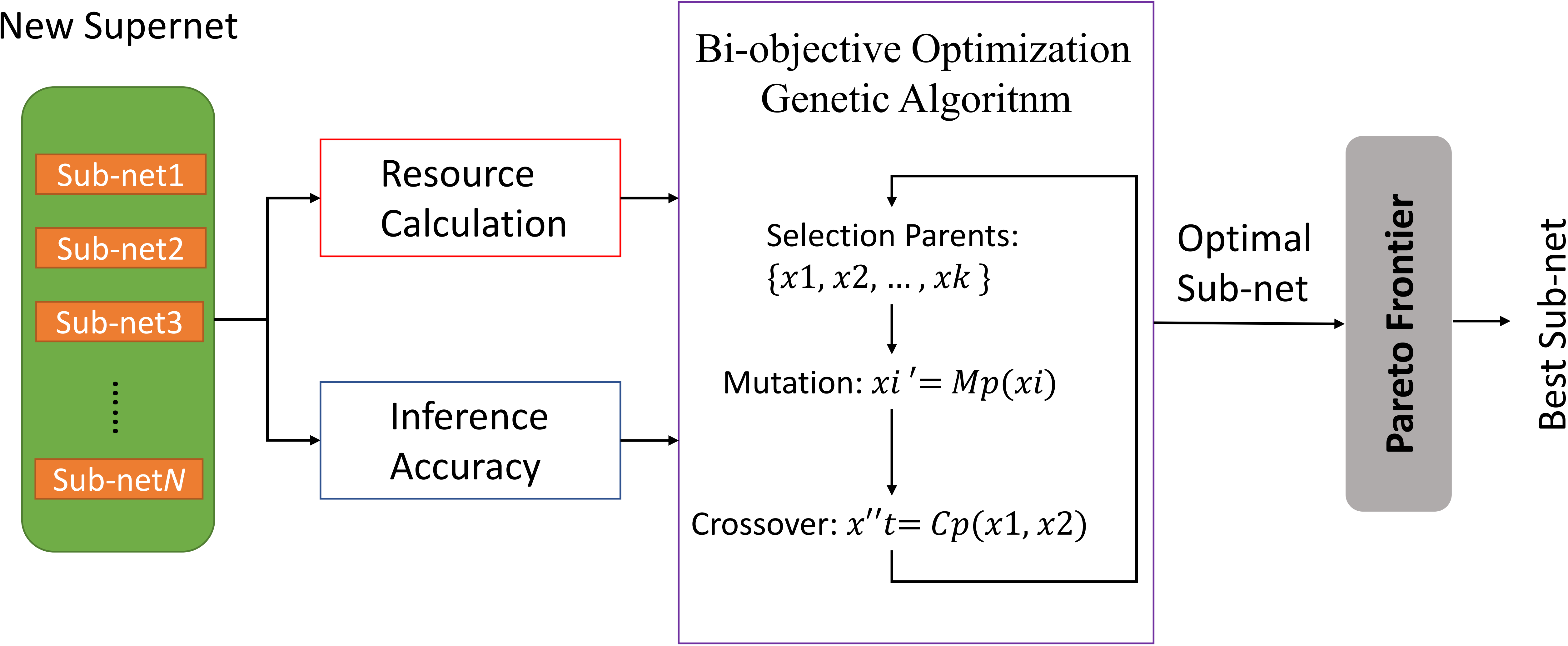}
    \caption{ Optimal Strategy Search}
    \label{fig7c}
  \end{subfigure}
  \caption{A NAS-based method to search for the optimal slicing strategy.}
  \label{fig7}
\end{figure}

For the aforementioned slicing strategy, both the size of the sliced image patches and the size of the overlapping regions are uncertain, constituting different slicing schemes. Smaller patch sizes are more hardware-friendly for deployment, but they also result in greater loss of model accuracy. It is difficult to manually balance these two aspects and find the globally optimal solution. Therefore, this paper proposes a NAS-based method to search for the optimal slicing strategy. As shown in Fig.~\ref{fig7}, the procedure consists of three steps: supernet construction, fine-tuning, and optimal strategy search.

\subsection{Supernet Construction and Fine-Tuning}
To search for the optimal slicing strategy, we first construct a continuous search space. This search space contains all slicing schemes with different slicing sizes and overlapping sizes, and each slicing scheme corresponds to a sub-network (sub-net). These sub-nets collectively form a supernet. Specifically, the input image is divided into slices of size $H_S \times W_S$, where $H_S$ and $W_S$ can take $Num_H$ and $Num_W$ possible values, respectively. This creates a continuous search space containing $Num_H \times Num_W$ sub-nets. Furthermore, we introduce three sizes of overlapping: 0 pixel, 1 pixel, and 2 pixels. With the introduction of overlapping, the search space expands to include $3 \times Num_H \times Num_W$ sub-nets.
Independently training each sub-net will incur significant costs. Therefore, in this work, all sub-nets share the same weight parameters. In other words, the slicing operation is only involved during inference and not during training.

Since all sub-nets share the same weights, it is only necessary to train a single supernet. After the supernet is trained, a fine-tuning process is required. The standard teacher-student paradigm typically uses \textit{KL}-divergence to measure the discrepancy between the teacher and student networks. However, for a supernet, \textit{KL}-divergence may fail to cover one or more modes of the teacher model, leading to severe penalties on the student model. Therefore, inspired by AlphaNet\cite{12}, this work adopts $\alpha$-divergence for fine-tuning the supernet:

\begin{equation}
\begin{aligned}
&D_{\alpha +, \alpha -} (p \| q) = \max \left\{ D_{\alpha +} (p \| q), D_{\alpha -} (p \| q) \right\}, \\ &\text{where } 
D_{\alpha} (p \| q) = \frac{1}{\alpha (\alpha - 1)} \sum_{i=1}^{m} q_i \left[ \left( \frac{p_i}{q_i} \right)^{\alpha} - 1 \right].
\end{aligned}
\end{equation}

Finally, the fine-tuning process of the supernet adopts the adaptive-KD loss within the $\alpha$-divergence constraint, which is formulated as:
\begin{equation}
    \begin{aligned}
\mathcal{L}_{\mathrm{KD}}(\theta, A) = \mathbb{E}_{x \sim D} 
& \left[ D_{\alpha +, \alpha -} \left( p(x; \theta, \alpha_b) \,\|\, q(x; \theta, A) \right) \right], \\
& \text{where } A \in \{\alpha_s, \alpha_r\}\,.
\end{aligned}
\end{equation}

\subsection{Optimal Strategy Search}
After the construction and training of the supernet are completed, we search for the optimal sub-net within it. Two key metrics are first defined for the search: 

\subsubsection{Resource Consumption}
The computational formula for resource consumption $R$ is defined as follows:

\begin{equation}
    R = \text{BitWidth} \times (W_S + O_W) \times (H_S + O_H) + \beta,
\end{equation}

\noindent where $W_S$ and $H_S$ are the slice width and height, respectively, and $O$ represents the size of the overlapping region. The parameter $\beta$ is a constant indicating the resources consumed by other components of the hardware system.

\subsubsection{Accuracy}
The accuracy metric is defined as the inference accuracy of the sub-net.

Since the search for the optimal strategy involves balancing these two metrics, it can be formulated as a bi-objective optimization problem. As shown in Algorithm~\ref{alg:boea}, our search employs a sampling-based genetic algorithm, where the search process constitutes a multi-objective genetic algorithm optimization with two objectives corresponding to the aforementioned search metrics. Specifically, each iteration of the search algorithm consists of four steps: selection, updating the Pareto frontier, mutation, and crossover. Mutation randomly modifies the selected slicing strategy, while crossover generates new slicing strategies by recombining two existing ones.

\begin{algorithm}[H]
\caption{Bi-Objective Evolutionary Search for Neural Sub-nets}
\label{alg:boea}
\begin{algorithmic}[1]
\STATE \textbf{Initialize:}
\STATE \quad Search space of sub-net architectures $\mathcal{X}$.
\STATE \quad Objective functions: $F_1(x) = \text{Acc}(x)$, $F_2(x) = \text{Resource}(x)$.
\STATE \quad Resource bounds: $R_{\min}$ (min resource), $R_{\max}$ (max resource).

\FOR{$t = 1$ to $T$}
    \STATE \textbf{Selection:}
    \STATE \quad Randomly sample $k$ sub-nets $\{x_1, \dots, x_k\} \subset \mathcal{X}$.
    \STATE \quad Evaluate objectives:
    \[
    F_1(x_i) = \text{Acc}(x_i), \quad F_2(x_i) = \text{Resource}(x_i) \quad \forall x_i
    \]

    \STATE \textbf{Update Pareto Front $\mathcal{P}$:}
    \FOR{each sub-net $x_i$}
        \IF{$F_2(x_i) \in [R_{\min}, R_{\max}]$}
            \STATE Add $x_i$ to $\mathcal{P}$ if non-dominated.
        \ENDIF
    \ENDFOR

    \STATE \textbf{Mutation:}
    \STATE \quad Randomly pick $x_i \in \mathcal{P}$, perturb its parameters:
    \[
    x_i' = x_i + \epsilon, \quad \epsilon \sim \mathcal Random
    \]
    \STATE \quad Clip mutated $W', H'$ to valid ranges.

    \STATE \textbf{Crossover:}
    \STATE \quad Select $x_a, x_b \in \mathcal{P}$, perform crossover:
    \[
    x_i'' = C_p(x_a,x_b),
    \]
    \STATE \quad where the function $C_p$ performs crossover on the $H_S$ (height) and $W_S$ (width) values of two sub-net slices with probability $P$, generating new sub-net parameters.
\ENDFOR
\STATE \textbf{Output:} Pareto-optimal sub-nets $\mathcal{P}$.
\end{algorithmic}
\end{algorithm}

\section{Experiment Results}

\subsection{Experimental Setup}
All algorithmic experiments are conducted on NVIDIA A100 GPUs, with models and training procedures implemented using PyTorch. The experimental task was image classification using the ImageNet-1K dataset~\cite{13},and the pre-trained model employed in our experiments is DAT~\cite{7}. During both the training and inference phases, the input image dimensions ($W$ and $H$) are consistently set to 224$\times$224 pixels. For our algorithmic experiments, we utilized the DAT-based model for both training and inference.



\subsection{Fine-Tuning Performance of the Supernet}
After the supernet is constructed, it undergoes five epochs of fine-tuning. The fine-tuning results are presented in Table~\ref{tab:accuracy_changes}. Epoch 0 refers to the original accuracy of DAT without any fine-tuning. Through this process, our model achieves a 0.2\% improvement in model accuracy.

\begin{table}[h]
\centering
\caption{Top-1 Accuracy during Fine-Tuning after Slicing}
\label{tab:accuracy_changes}
\begin{tabular}{cc}
\toprule
\textbf{Epoch of fine-tuning} & \textbf{Accuracy} \\
\midrule
0 & 84.6\% \\
1 & 84.6\% \\
2 & 84.7\% \\
3 & 84.7\% \\
4 & 84.7\% \\
\bottomrule
\end{tabular}
\end{table}

To investigate the impact of our hardware-friendly optimizations on model accuracy, we conduct an ablation study as detailed as shown in Table~\ref{tab:ablation}. The results demonstrate that our optimizations result in only a 0.2\% accuracy drop, while this marginal degradation could be effectively compensated through the aforementioned supernet fine-tuning procedure.

\begin{table}[htbp]
\centering
\caption{Ablation Experiment Results of the Algorithm}
\label{tab:ablation}
\begin{tabular}{cccc|c}
\toprule
\multicolumn{4}{c|}{\textbf{Method}} & \textbf{Top-1} \\
\cmidrule(lr){1-4}
\textbf{Slicing} & \textbf{Overlap} & \textbf{Fine-} & & \textbf{Accuracy} \\
\textbf{Strategy} & \textbf{Slicing} & \textbf{Tuning} & & \\
\midrule
\textcolor{red}{\ding{55}} & \textcolor{red}{\ding{55}} & \textcolor{red}{\ding{55}} & & 84.9\% \\
\textcolor{green!50!black}{\checkmark} & \textcolor{red}{\ding{55}} & \textcolor{red}{\ding{55}} & & 84.6\% \\
\textcolor{green!50!black}{\checkmark} & \textcolor{green!50!black}{\checkmark} & \textcolor{red}{\ding{55}} & & 84.6\% \\
\textcolor{green!50!black}{\checkmark} & \textcolor{green!50!black}{\checkmark} & \textcolor{green!50!black}{\checkmark} & & 84.7\% \\
\bottomrule
\end{tabular}
\end{table}

\subsection{Algorithmic Result of Optimal Slicing Strategy}
After finishing the supernet fine-tuning, the proposed NAS-based method identifies the optimal slicing strategy for the given input image and model architecture. In our study, the input image size of the deformable attention layer is 56$\times$56, and the slicing size is limited to less than or equal to 28. In addition, the slicing size should be greater than the 7$\times$7 window size. Therefore, the values of $H_S$ and $W_S$ range between 8 and 28, and the best slicing values searched by our method are: $W_s =14$ and $H_s =28$.

We compared the top-1 accuracy of other works that also performed classification tasks on the ImageNet-1K dataset. Our method achieved an accuracy of 84.7\%, outperforming similar approaches as shown in Table~\ref{tab:nas_results}.

\begin{table}[h]
\centering
\caption{Top-1 Accuracy Comparison with other works}
\label{tab:nas_results}
\begin{tabular}{ccc}
\toprule
\textbf{Method} & \textbf{Dataset} & \textbf{Accuracy} \\
\midrule
MambaVision-B~\cite{mambavision} & ImageNet-1K & 84.1\% \\
\midrule
GroupMamba~\cite{groupmamba} & ImageNet-1K & 83.9\% \\
\midrule
 SpectFormer~\cite{spectformer} & ImageNet-1K & 80.21\% \\
\midrule
Our Work & ImageNet-1K & 84.7\% \\
\bottomrule
\end{tabular}
\end{table}


\begin{figure}[htb]
  \centering
  \includegraphics[width=0.95\linewidth]{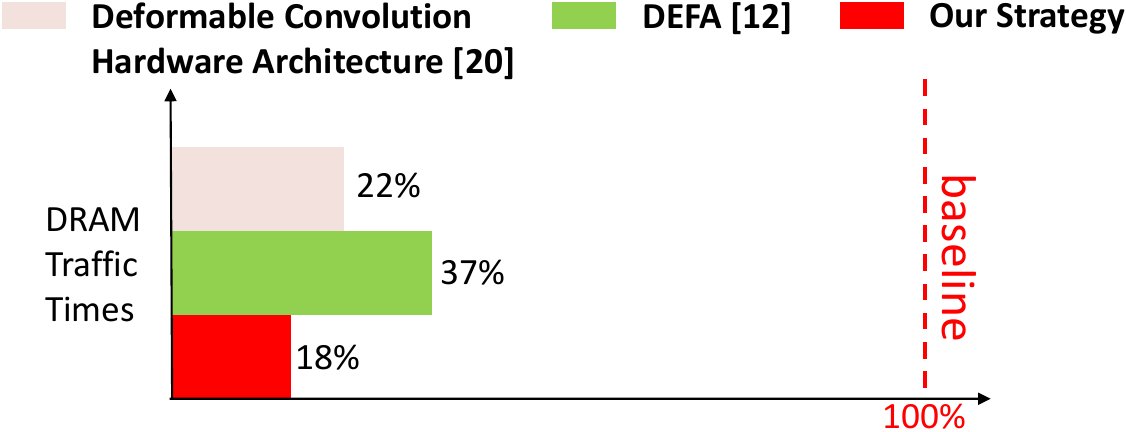}
  \caption{Normalized DRAM traffic times comparisons.}
  \label{fig-7}
\end{figure}
\subsection{Hardware Overhead Analysis}
To verify the advantages of our framework in terms of hardware deployment, we implement the optimized DAT on the Xinlin FPGA platform using Vivado 2019.1 with verilog HDL, the hardware architecture is designed refer to~\cite{SOCC2024Zen}. For a fair comparison,  we normalize the DRAM access times for the deformable attention/convolutional layer, following the approach proposed in~\cite{14}, and normalize times for the methods of DEFA~\cite{10} and~\cite{deCNN}. As shown in Fig.~\ref{fig-7}, our method decreases DRAM access times to 18\% compared with baseline (layer by layer processing the sampling layer and the attention layer),  significantly reducing bandwidth resources and power consumption. DEFA utilizes operator fusion and feature map reuse techniques to improve hardware efficiency. In addition, it develops multi-scale grid-sampling scheme(MSGS), reducing DRAM access times to nearly 37\%. However, MSGS adopts the frequency-weighted pruning method to Optimize memory access, which added hardware overhead for masks and control logic. Our method proposes a hardware-friendly framework without modifying algorithm architecture, significantly reducing the hardware costs in terms of memory access and logical resources.



\section{Conclusion}
This paper presents a comprehensive framework for accelerating deformable attention mechanisms. By introducing a training-free slicing strategy, we effectively address the irregular memory access challenges due to data-dependent sampling, enabling parallel processing and reduced hardware resource consumption. Using a memory-aware NAS algorithm, our method automatically identifies optimal slice configurations that balance hardware efficiency and model performance. The experimental results demonstrate the dual advantages of our method in terms of algorithm accuracy and hardware efficiency.

\bibliographystyle{IEEEtran}
\bibliography{reference}

\begin{thebibliography}{10}
\providecommand{\url}[1]{#1}
\csname url@samestyle\endcsname
\providecommand{\newblock}{\relax}
\providecommand{\bibinfo}[2]{#2}
\providecommand{\BIBentrySTDinterwordspacing}{\spaceskip=0pt\relax}
\providecommand{\BIBentryALTinterwordstretchfactor}{4}
\providecommand{\BIBentryALTinterwordspacing}{\spaceskip=\fontdimen2\font plus
\BIBentryALTinterwordstretchfactor\fontdimen3\font minus \fontdimen4\font\relax}
\providecommand{\BIBforeignlanguage}[2]{{%
\expandafter\ifx\csname l@#1\endcsname\relax
\typeout{** WARNING: IEEEtran.bst: No hyphenation pattern has been}%
\typeout{** loaded for the language `#1'. Using the pattern for}%
\typeout{** the default language instead.}%
\else
\language=\csname l@#1\endcsname
\fi
#2}}
\providecommand{\BIBdecl}{\relax}
\BIBdecl

\bibitem{1}
A.~Vaswani, N.~Shazeer, N.~Parmar, J.~Uszkoreit, L.~Jones, A.~N. Gomez, L.~Kaiser, and I.~Polosukhin, ``Attention is all you need,'' in \emph{Proceedings of the 31st International Conference on Neural Information Processing Systems}.\hskip 1em plus 0.5em minus 0.4em\relax Red Hook, NY, USA: Curran Associates Inc., 2017, p. 6000–6010.

\bibitem{2}
A.~Dosovitskiy, L.~Beyer, A.~Kolesnikov, D.~Weissenborn, X.~Zhai, T.~Unterthiner, M.~Dehghani, M.~Minderer, G.~Heigold, and S.~a. Gelly, ``An image is worth 16x16 words: Transformers for image recognition at scale,'' in \emph{International Conference on Learning Representations}, 2021.

\bibitem{3}
N.~Carion, F.~Massa, G.~Synnaeve, N.~Usunier, A.~Kirillov, and S.~Zagoruyko, ``End-to-end object detection with transformers,'' in \emph{Computer Vision -- ECCV 2020}, A.~Vedaldi, H.~Bischof, T.~Brox, and J.-M. Frahm, Eds.\hskip 1em plus 0.5em minus 0.4em\relax Cham: Springer International Publishing, 2020, pp. 213--229.

\bibitem{4}
\BIBentryALTinterwordspacing
H.~Touvron, M.~Cord, M.~Douze, F.~Massa, A.~Sablayrolles, and H.~Jegou, ``Training data-efficient image transformers \& amp; distillation through attention,'' in \emph{Proceedings of the 38th International Conference on Machine Learning}, ser. Proceedings of Machine Learning Research, M.~Meila and T.~Zhang, Eds., vol. 139.\hskip 1em plus 0.5em minus 0.4em\relax PMLR, 18--24 Jul 2021, pp. 10\,347--10\,357. [Online]. Available: \url{https://proceedings.mlr.press/v139/touvron21a.html}
\BIBentrySTDinterwordspacing

\bibitem{5}
H.~Wang, Y.~Zhu, B.~Green, H.~Adam, A.~Yuille, and L.-C. Chen, ``Axial-deeplab: Stand-alone axial-attention for panoptic segmentation,'' in \emph{Computer Vision -- ECCV 2020}, A.~Vedaldi, H.~Bischof, T.~Brox, and J.-M. Frahm, Eds.\hskip 1em plus 0.5em minus 0.4em\relax Cham: Springer International Publishing, 2020, pp. 108--126.

\bibitem{6}
Z.~Liu, Y.~Lin, Y.~Cao, H.~Hu, Y.~Wei, Z.~Zhang, S.~Lin, and B.~Guo, ``Swin transformer: Hierarchical vision transformer using shifted windows,'' in \emph{2021 IEEE/CVF International Conference on Computer Vision (ICCV)}, 2021, pp. 9992--10\,002.

\bibitem{Mobilevit}
S.~Mehta and M.~Rastegari, ``{MobileViT}: light-weight, general-purpose, and mobile-friendly vision transformer,'' in \emph{International Conference on Learning Representations (ICLR)}, 2022.

\bibitem{LeViT}
B.~Graham \emph{et~al.}, ``{LeViT}: a vision {Transformer} in convnet's clothing for faster inference,'' in \emph{2021 IEEE/CVF International Conference on Computer Vision (ICCV)}, 2021, pp. 12\,259--12\,269.

\bibitem{7}
Z.~Xia, X.~Pan, S.~Song, L.~E. Li, and G.~Huang, ``Vision transformer with deformable attention,'' in \emph{2022 IEEE/CVF Conference on Computer Vision and Pattern Recognition (CVPR)}, 2022, pp. 4784--4793.

\bibitem{8}
H.~Wang, Z.~Zhang, and S.~Han, ``Spatten: Efficient sparse attention architecture with cascade token and head pruning,'' in \emph{2021 IEEE International Symposium on High-Performance Computer Architecture (HPCA)}, 2021, pp. 97--110.

\bibitem{9}
T.~J. Ham, Y.~Lee, S.~H. Seo, S.~Kim, H.~Choi, S.~J. Jung, and J.~W. Lee, ``Elsa: Hardware-software co-design for efficient, lightweight self-attention mechanism in neural networks,'' in \emph{2021 ACM/IEEE 48th Annual International Symposium on Computer Architecture (ISCA)}, 2021, pp. 692--705.

\bibitem{10}
Y.~Xu, D.~Lyu, Z.~Li, Z.~Wang, Y.~Chen, G.~Wang, Z.~Wang, H.~Li, and G.~He, ``Defa: Efficient deformable attention acceleration via pruning-assisted grid-sampling and multi-scale parallel processing,'' \emph{arXiv preprint arXiv:2403.10913}, 2024.

\bibitem{12}
\BIBentryALTinterwordspacing
D.~Wang, C.~Gong, M.~Li, Q.~Liu, and V.~Chandra, ``Alphanet: Improved training of supernets with alpha-divergence,'' in \emph{Proceedings of the 38th International Conference on Machine Learning}, ser. Proceedings of Machine Learning Research, M.~Meila and T.~Zhang, Eds., vol. 139.\hskip 1em plus 0.5em minus 0.4em\relax PMLR, 18--24 Jul 2021, pp. 10\,760--10\,771. [Online]. Available: \url{https://proceedings.mlr.press/v139/wang21i.html}
\BIBentrySTDinterwordspacing

\bibitem{13}
\BIBentryALTinterwordspacing
O.~Russakovsky, J.~Deng, H.~Su, J.~Krause, S.~Satheesh, S.~Ma, Z.~Huang, A.~Karpathy, A.~Khosla, M.~Bernstein, A.~C. Berg, and L.~Fei-Fei, ``The imagenet large scale visual recognition challenge,'' \emph{International Journal of Computer Vision (IJCV)}, vol. 115, no.~3, pp. 211--252, 2015. [Online]. Available: \url{https://link.springer.com/article/10.1007/s11263-015-0816-y}
\BIBentrySTDinterwordspacing

\bibitem{mambavision}
A.~Hatamizadeh and J.~Kautz, ``Mambavision: A hybrid mamba-transformer vision backbone,'' in \emph{Proceedings of the Computer Vision and Pattern Recognition Conference}, 2025, pp. 25\,261--25\,270.

\bibitem{groupmamba}
A.~Shaker, S.~T. Wasim, S.~Khan, J.~Gall, and F.~S. Khan, ``Groupmamba: Efficient group-based visual state space model,'' in \emph{Proceedings of the Computer Vision and Pattern Recognition Conference (CVPR)}, June 2025, pp. 14\,912--14\,922.

\bibitem{spectformer}
B.~N. Patro, V.~P. Namboodiri, and V.~S. Agneeswaran, ``Spectformer: Frequency and attention is what you need in a vision transformer,'' in \emph{2025 IEEE/CVF Winter Conference on Applications of Computer Vision (WACV)}, 2025, pp. 9543--9554.

\bibitem{SOCC2024Zen}
Q.~Zeng, Y.~Wang, Z.~Wang, and W.~Mao, ``An automated hardware design framework for various dnns based on chatgpt,'' in \emph{2024 IEEE 37th International System-on-Chip Conference (SOCC)}, 2024, pp. 1--6.

\bibitem{14}
F.~Tu, S.~Yin, P.~Ouyang, S.~Tang, L.~Liu, and S.~Wei, ``Deep convolutional neural network architecture with reconfigurable computation patterns,'' \emph{IEEE Transactions on Very Large Scale Integration (VLSI) Systems}, vol.~25, no.~8, pp. 2220--2233, 2017.

\bibitem{deCNN}
Y.~Yu, J.~Luo, W.~Mao, and Z.~Wang, ``A memory-efficient hardware architecture for deformable convolutional networks,'' in \emph{2021 IEEE Workshop on Signal Processing Systems (SiPS)}, 2021, pp. 140--145.

\end{thebibliography}

\end{document}